\documentclass[sigconf]{acmart}
\usepackage{booktabs} % For formal tables
\usepackage{xcolor} % For Color

\usepackage[linesnumbered,ruled]{algorithm2e}

\DeclareMathOperator*{\argmax}{argmax}

\newcommand*{\argmaxl}{\argmax\limits}

\setcopyright{none}
\renewcommand\footnotetextcopyrightpermission[1]{} % removes footnote with confere
\copyrightyear{2019} 
\acmYear{2019} 
\setcopyright{acmlicensed}
\acmConference[ICMR '19]{International Conference on Multimedia Retrieval}{June 10--13, 2019}{Ottawa, ON, Canada}
\acmBooktitle{International Conference on Multimedia Retrieval (ICMR '19), June 10--13, 2019, Ottawa, ON, Canada}
\acmPrice{15.00}
\acmDOI{10.1145/3323873.3325052}
\acmISBN{978-1-4503-6765-3/19/06}

\begin{document}
%\title{Adversarial Image Queries: Blocking Retrieval by Modifying Neural, Local and Global Features}
%\title{Adversarial Queries: Blocking Content-based Image Retrieval by Image Features}
\title[Who's Afraid of Adversarial Queries?]{Who's Afraid of Adversarial Queries? The Impact of Image Modifications on Content-based Image Retrieval}
%\titlenote{Produces the permission block, and
%  copyright information}
%\subtitle{}
% \subtitlenote{The full version of the author's guide is available as
%   \texttt{acmart.pdf} document}

\author{Zhuoran Liu, Zhengyu Zhao, Martha Larson}
\affiliation{Radboud University, Netherlands}
\email{\string{z.liu, z.zhao, m.larson\string}@cs.ru.nl}

\begin{abstract}
An adversarial query is an image that has been modified to disrupt content-based image retrieval (CBIR), while appearing nearly untouched to the human eye. 
This paper presents an analysis of adversarial queries for CBIR based on neural, local, and global features.
We introduce an innovative neural image perturbation approach, called Perturbations for Image Retrieval Error (PIRE), that is capable of blocking neural-feature-based CBIR.
%To our knowledge PIRE is the first approach to creating neural adversarial examples for CBIR.
PIRE differs significantly from existing approaches that create images adversarial with respect to CNN classifiers because it is unsupervised, i.e., it needs no labeled data from the data set to which it is applied.
Our experimental analysis demonstrates the surprising effectiveness of PIRE in blocking CBIR, and also covers aspects of PIRE that must be taken into account in practical settings, including saving images, image quality and leaking adversarial queries into the background collection.
Our experiments also compare PIRE (a neural approach) with existing keypoint removal and injection approaches (which modify local features).
Finally, we discuss the challenges that face multimedia researchers in the future study of adversarial queries.
\end{abstract}



\keywords{Content-based Image Retrieval, Adversarial Examples, Deep Learning, Local Image Features, SIFT, Global Image Features, Image Forensics, Privacy}

\maketitle

\section{Introduction}
\label{sec:intro}
%What does this paper do?
%Adversarial examples are common in classification, but no one has looked at them for IR.
Recently, researchers working on deep learning for image classification have started to study adversarial images intensively and to develop techniques to create them~\cite{szegedy2013intriguing}~\cite{43405}~\cite{moosavi2016deepfool}~\cite{carlini2017towards}~\cite{Moosavi-Dezfooli_2017_CVPR}~\cite{mopuri-bmvc-2017}. 
%check citation scope, add more references
Their work defines an \emph{adversarial example} to be an image that a human can easily interpret, but that a CNN-based classifier assigns to an unexpected class. 
Typically, adversarial examples are created by taking an image that is correctly classified by a classifier, and perturbing the pixels. 
%cite
The perturbations are small, such that humans can look at the modified image and judge it to be nearly untouched. 
The perturbations are also carefully chosen, such that the modified image is no longer classified correctly, but rather is moved over the decision boundary of the classifier and is classified incorrectly.
Generally, an image set labeled with the target classes is used to train the perturbations.

\begin{figure}
\includegraphics[width=1.0\columnwidth]{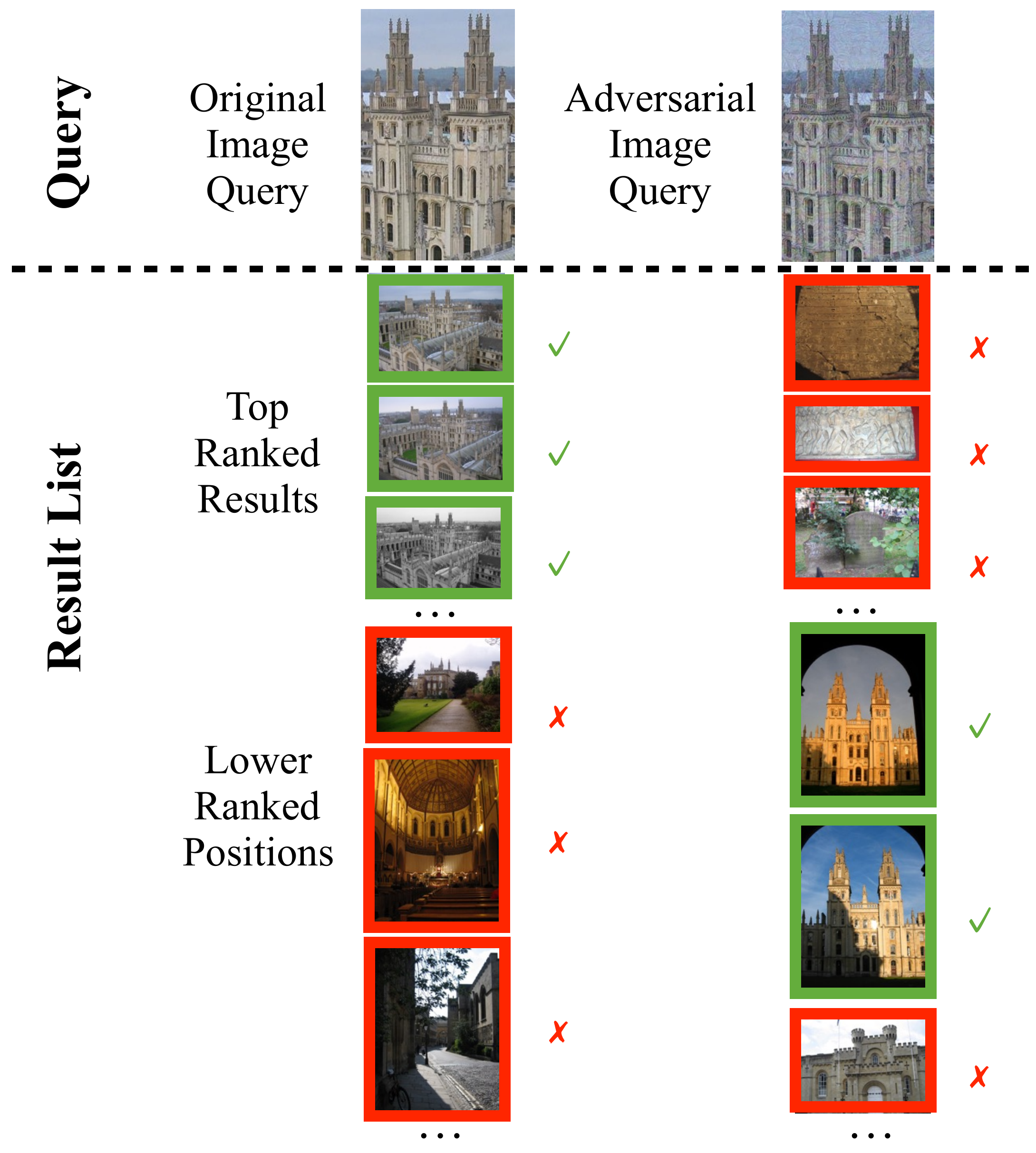}
\vspace{-0.3cm}
\caption{A successful query image (top left) and the corresponding adversarial query (top right). The two are visually nearly identical to the human eye. A CBIR system ranks relevant results high for the original image query and low for the adversarial query.}
\label{fig:dl-ex1}
\vspace{-0.4cm}
\end{figure}

%What do we do in this paper?
In this paper, we extend the idea of adversarial examples from image classification to content-based image retrieval (CBIR). 
We define an \emph{adversarial query} as an image that a human can easily interpret, but that causes a CBIR system unexpected difficulties. 
The principle is illustrated by the example in Figure~\ref{fig:dl-ex1}.
The adversarial query resembles the original image as closely as possible.\footnote{Our code is available at \url{https://github.com/liuzrcc/PIRE}} 

%What is the difference between adversarial queries and adversarial examples? What is our main contribution?
The fundamental difference between creating
adversarial examples in the case of image classification and
creating adversarial queries in the case of CBIR is the information available for guiding the image modifications. 
In contrast to classification systems, which assume a set of discrete classes, CBIR systems are designed to handle arbitrary queries and unconstrained background collections.
Specifically, in the deep learning image classification scenario, adversarial modifications are informed by class boundaries.
Decision boundary information is lacking in the CBIR scenario. 
As such, in the CBIR scenario, there is no obvious direction, or directions, in which to move an image in pixel space in order to create a query that is adversarial with respect to the CBIR system.
In order to address this challenge, we propose an approach called Perturbations for Image Retrieval Error (PIRE). 
%In order to create queries that are adversarial with respect to a CBIR system that uses neural representations, 

%As such, there is no obvious information about which direction an image must be moved in pixel space in order to train an approach that can create queries that are adversarial with respect to a CBIR system that uses neural representations. 
%We propose an approach, called Perturbations for Image Retrieval Error (PIRE), that is able to generate perturbations without needing such information (i.e., PIRE requires no class labels, or relevance judgements from the data set to which it is applied). 
PIRE is able to generate perturbations without needing guiding information (i.e., PIRE requires no class labels, or relevance judgments from the data set to which it is applied).
PIRE perturbs images such that they can still be interpreted to the human eye, but that they no longer can be used as successful queries for CBIR.

In sum, this paper makes the following contributions: (1) We explain why it is important to study adversarial queries. (2) We present PIRE, our neural perturbation approach to creating adversarial queries, and experimentally demonstrate its impact on different CBIR systems (i.e., systems using neural, local, and global features). (3) We discuss and analyze practical aspects of adversarial queries.
%, focusing on issues that risk being overlooked.
%It is important to note that we do not claim to provide the ultimate solution for creating queries that are adversarial with respect to any arbitrary CBIR scenario.
%Rather, our experiments and discussion are directed at providing a deeper understanding of the potential and limitations of adversarial queries in order to lay a road map for future study.
The paper is organized as follows: after introducing the importance of adversarial queries in Section~\ref{sec:why}, we present the relevant related work in Section~\ref{sec:related_work}. 
Section~\ref{sec:exp_fra} describes the framework in which we carry out our experiments. Then, Sections~\ref{Neural_CBIR} and~\ref{sec:CBIR_beyond} present our experiments and analyses. Finally, Section~\ref{sec:conclusion} pulls everything together, and provides an outlook on future work.

\section{Why study Adversarial Queries?}
\label{sec:why}
The study of adversarial image examples is motivated by specific \emph{threat models}.
Informally defined, a threat model expresses what we should be worried about, i.e., the dangers that a specific system or technology must be able to ward off. 
Adversarial image queries play a role in widely different threat models, which are described in this section.
%This section describes those threat models.
%, showing that adversarial queries can be approached from a multiplicity of perspectives.
Section~\ref{sec:threat_modi} looks at adversarial queries as being dangerous. 
From this perspective, the practical application of our research is understanding attacks on CBIR systems in order to defend against them.
Section~\ref{sec:threat_ir} looks at adversarial queries as being protective. 
From this perspective, the practical application of our research is preventing, or at least disincentivizing, harmful use of CBIR.

%The importance of different perspectives on adversarial queries inspired us to include a question in the title of our paper. 
%We return to this question in the final section.

\subsection{Threat of image modification technology}
\label{sec:threat_modi}
The assumption behind many widely-adopted threat models is that modified images are a source of danger. 
Here, we discuss three familiar examples of such threat models.
First, researchers working on image classification are generally worried about scenarios in which modified images cause misclassification.
This threat model applies, for example, to scenarios in which computer vision technology is used by self-driving cars~\cite{Eykholt_2018_CVPR}.
Adversarial image queries are relevant to such threat models since memory-based image classifiers are generally based on CBIR systems. 
Second, researchers working on keypoint removal and injection are generally worried about scenarios in which modified images cause the identification of duplicate or near-duplicate images to be blocked.
Such work, e.g.,~\cite{do2010understanding},~\cite{do2010deluding} and ~\cite{do2012enlarging}, is carried out in the use scenario of preventing copyright violation and image forgery via copy-move. 
Third, researchers working on image forensics also care about the post-processing operations, such as resampling~\cite{popescu2005exposing}, double JPEG compression~\cite{fridrich2008detection} and denoising~\cite{kirchner2010detection}, since they are of interest in a forensic examination of an image and can affect forensic methods in various ways~\cite{kirchner2010detection}.
% Here, we have mentioned two examples, but we point out that literature on forensics %Alessandro Piva, “An Overview on Image Forensics,” ISRN Signal Processing, vol. 2013, Article ID 496701, 22 pages, 2013. https://doi.org/10.1155/2013/496701.
% readily reveals others.
If we consider these threat models, then our reason for studying adversarial queries is to understand how modified images can harm image matching systems.

\subsection{Threat of image retrieval technology}
\label{sec:threat_ir}
The assumption behind another more recently emerging class of threat models is that the multimedia retrieval system itself is a source of danger.
These retrieval-specific threat models are commonly adopted by researchers working on multimedia privacy.
The specific threat is a privacy violation, specifically, harm that people suffer caused by malicious actors who misuse an existing retrieval system (for example an online image search engine) or who build their own retrieval system to search in a collection of misappropriated images.
This danger was first articulated by ~\cite{friedland2010cybercasing}, who described the threat of `cybercasing': criminals using online search engines to mine the Web for users whose online sharing behavior reveals that they own valuable items, and when they are away on vacation.
The concern has been recently grown stronger because of high profile data breaches, e.g.,~\cite{weried18Dec}, which have made clear that sharing images in `private' mode is not a perfect solution for protection.
Unscrupulous actors can implement their own CBIR system if they can get their hands on enough data.
The interesting and surprising aspect of retrieval-specific threat models is that giving people access to image modification technology actually would help them to protect themselves against those seeking to misuse their images.
Instead of a danger, image modification is a form of protection.
If we consider retrieval-specific threat models, then our reason for studying adversarial queries is to understand the conditions under which the matching ability of CBIR systems can be blocked.

A recent investigation concerned with the threat of cybercasing~\cite{choi2017geo} examines the potential of image enhancements to block the inference of the geo-location of the photos that users take and post online. 
Our work differs from~\cite{choi2017geo} in that we focus specifically on CBIR and we consider image queries that are explicitly designed to be adversarial.
However, we adopt the same threat model, and we focus our investigation on a CBIR problem that is related to location because it involves images of buildings in cities.

\section{Related Work}
\label{sec:related_work}
We first cover work on neural adversarial examples for image classification and then work on blocking local-feature-based CBIR.

\subsection{Adversarial examples and  classification}
Research on adversarial examples in the deep learning community was launched by~\cite{szegedy2013intriguing}, who demonstrated the possibility of constructing images adversarial with respect to a convolutional neural network image classifier (which we will refer to as the 'CNN-model').
%To generate adversarial examples, an optimization task is defined to change the predicted label of perturbed image, while minimizing the distance between perturbed image and original image.
As mentioned in the introduction, the basic mechanism used to create adversarial examples is to perturb pixels to construct a misclassified image while at the same time minimizing the distance between the original image (input image) and adversarial image.
Work on adversarial examples started with `whitebox' approaches, which have full knowledge of the CNN-model that they are attempting to delude.
The Fast Gradient Sign Method (FGSM)~\cite{43405} makes use of the gradient of the model with respect to the input image.
It increases the model's loss on the input image given the correct class label by perturbing it in the ascending direction of the gradient. 
DeepFool~\cite{moosavi2016deepfool} extends FGSM with more precise control over the size of the perturbations. 
For both FGSM and DeepFool, the perturbations are specific to the input image, and the correct class label of that image is known. Comparatively, PIRE only operates on neural features without accessing any ground truth (e.g., relevance judgements) of the CBIR system.
%Discuss this point

Subsequently, researchers have worked to extend `whitebox' methods so that they require less information about the input images and less information about the CNN-model. 
Universal Adversarial Perturbations (UAP)~\cite{Moosavi-Dezfooli_2017_CVPR} took a first step in this direction. 
UAP produces perturbations that do not require prior knowledge of the input images, however it does need a labeled training set.
%perturbation which is unique and universal for given datasets.
UAP adversarial examples have been shown to have an adversarial effect on CNN-models other than the one originally used to generate the perturbations. 
The `universal' in UAP means that the perturbations are generated to be effective for a majority of images, although in practice they fail for a subset of images.
Another whitebox method that is universal in this respect is Fast Feature Fool (FFF)~\cite{mopuri-bmvc-2017}, which generates adversarial images by calculating the maximal spurious activations in each convolutional layer while constraining the size of perturbations.
FFF, like UAP, produces perturbations without knowledge of the images to be modified.
However, whereas UAP requires training data, FFF can make use of the CNN-model with no additional training needed.

%All methods above need exact information of neural networks and ground truth of dataset, e.g., gradients of input, input image labels and exact dataset, while in CBIR, exact annotated images may not available.

Currently, `blackbox' techniques, which can create images adversarial to an arbitrary CNN-model remain elusive.
Attempts at `blackbox' solutions leverage existing `whitebox' solutions.
An ensemble method has been proposed, which creates examples that are adversarial with respect to a number of known CNN-models, and then tests them against a blackbox model~\cite{liu2016delving}. 
Also, reconstruction methods have been proposed, namely~\cite{Papernot:2017:PBA:3052973.3053009} and ~\cite{papernot2016transferability}, which probe the blackbox model with test examples, and then train substitute models that mimic the real model.

In our work, we focus on the case where we have access to the trained CNN-model used by the CBIR system at the moment at which we create our adversarial queries.
%neural network used to create the a that has access to the trained CNN-model used to create the neural representation used by the CBIR system.
%We anticipate that our method can be extended later to `blackbox' settings by strategies similar to those just discussed.
However, we point out that our approach is not a completely `whitebox' approach. Labeled training data is used to pre-train and fine-tune the CNN-model, but PIRE is ultimately applied to images from a third semantically related, but yet completely unseen, data set.
For this reason, we refer to our approach as unsupervised, and not requiring class labels.

\subsection{Keypoint Removal and Injection (KR\&I)}
\label{sec:krni}
In order to provide a complete picture of the behavior of adversarial queries, we consider not only neural features, but also local features. 
We focus on SIFT-based methods because they are representative of local-feature-based CBIR and also due to the rich literature on SIFT KR\&I.
%The literature has shown that the impact of KR\&I on the performance of SIFT feature-based CBIR.
The first work to consider influences of KR\&I in SIFT-based CBIR systems was 
~\cite{do2010understanding}, ~\cite{do2010deluding} and ~\cite{do2012enlarging}.
Here, blocking CBIR means blocking the retrieval of exact duplicate or near duplicate images.
In contrast, we are interested in blocking the retrieval of images containing the same subject matter as the query, without a specific focus on matching duplicates or near duplicates.

Other KR\&I work is not directly connected to CBIR, but focuses on image forensics and multimedia security.
With security issues in mind, ~\cite{hsu2009secure} proposed to modify SIFT features while simultaneously keeping image quality. 
Later the authors proposed an optimization-based approach~\cite{lu2012constraint}.
Combining multiple techniques, Classification-based Attack (CLBA)~\cite{costanzo2014forensic} proposed to use an iterative procedure to apply different methods on keypoints in different classes.
~\cite{li2016sift} proposed SIFT keypoint removal and injection methods which remove keypoints with minimized distortion on the processed image.
Recently,~\cite{li2017sift} proposed Removal via Directed Graph Construction (RDG) method to remove SIFT keypoints for colour images while maintaining high visual quality.

%Most of KR\&I methods only consider image forensics and multimedia security issues.
%Their performance in general CBIR systems has not been evaluated.

\section{Experimental Framework}
\label{sec:exp_fra}

\begin{figure}
\vspace{-0.5cm}
\includegraphics[width=1.0\columnwidth]{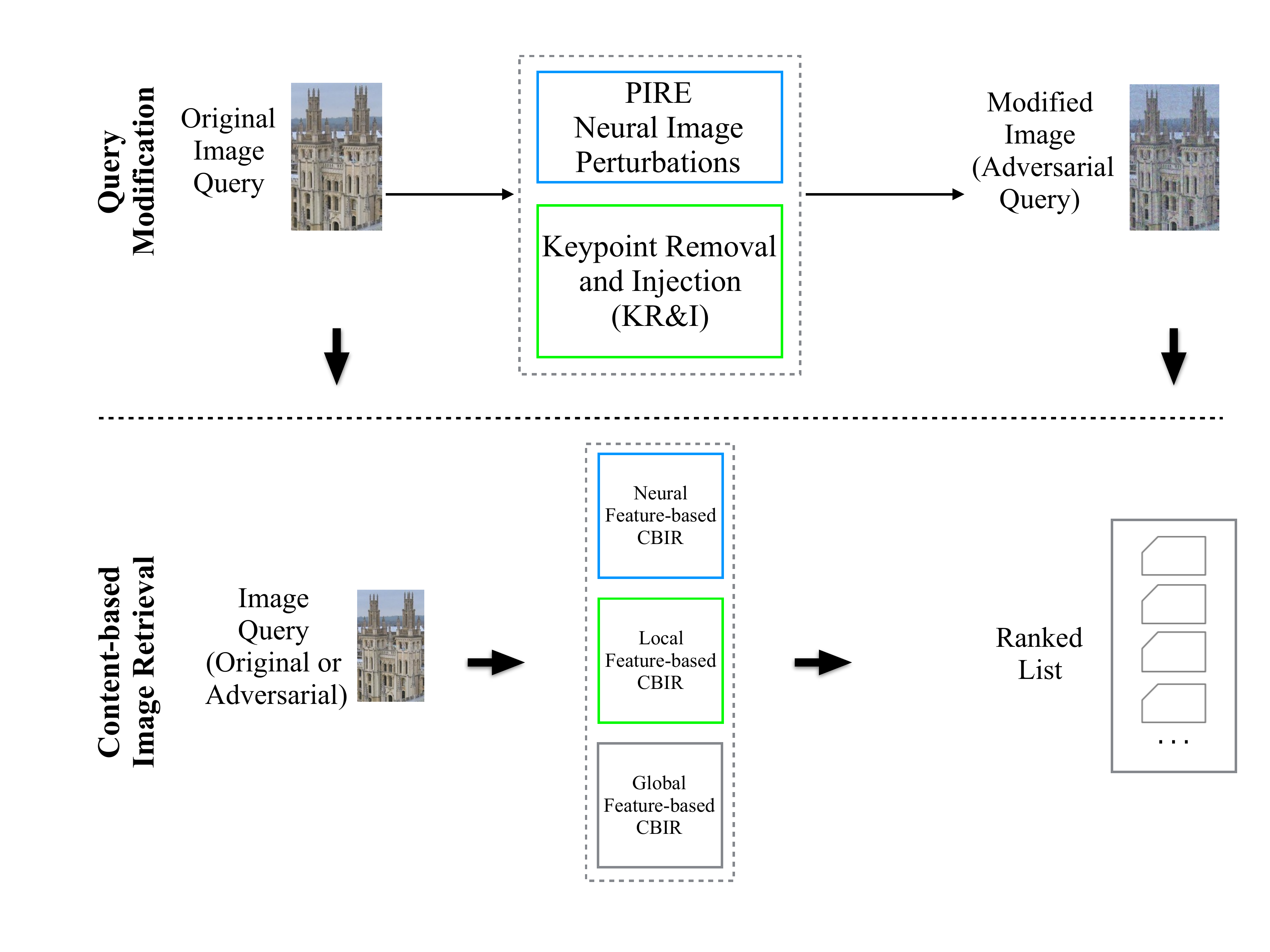}
\vspace{-0.8cm}
\caption{Our experimental analysis tests combinations of query modifications (Top) and CBIR systems (bottom). Blue boxes are neural-representation approaches and green boxes are local-feature approaches.}
%\caption{(Top) Framework of image modification with our PIRE approach and SIFT Keypoint Removal and Injection (KR\&I) approach ; (Bottom) Working diagram of the adversarial query image in three different types of CBIR systems, namely, neural features-based, local features-based and global features-based CBIR.}
\vspace{-0.5cm}
\label{fig:sys-flow}
\end{figure}

% We got figure two and go through the sessions

% In this section %In neural feature-based CBIR, 
Figure~\ref{fig:sys-flow} depicts the framework in which we carry out our experimental analysis.
Our experiments test different combinations of image modification approach and CBIR system.
The top of the figure shows the query modification step, which either uses PIRE or KR\&I.
The bottom of the figure shows the image retrieval step, which uses a CBIR system based on either neural, local, or global features.
In this section, we describe the design choices that we use for implementing the framework, before introducing PIRE in detail.
\subsection{Content-based Image Retrieval Systems}
\label{CBIR_system}
A CBIR system accepts an image as a query and returns a list of relevant images as a result.
The images are drawn from a larger collection, which we refer to as the background collection.
In a basic CBIR system, such as the one we adopt here, ranking occurs by comparing the vector representing the query image with the vectors representing each of the images in the background collection.
The results list consists of the images from the background collection ranked in order of closeness to the query.
CBIR systems are differentiated by the features that they use to create the image feature vectors.
As previously mentioned, in our experiments we use neural, local, and global features. 
We describe each in turn.

%According to different image representation (feature extraction) methods, CBIR systems are classified as neural-feature-based, local-feature-based and global- feature-based CBIR systems.

% CNN-model extracts image representations by extracting the output of last convolutional layer or fully connected layer.
Neural representations are compact representations that are extracted from an image using a pre-trained, and possibly then also fine-tuned, CNN-model.
%Given an image, a compact representation is extracted by a pre-trained (fine-tuned) CNN-model.
For our experiments, we need the currently best available neural representations, and for this reason we adopt GeM~\cite{Radenovic-TPAMI18}.
GeM is a fully convolutional CNN-model with a Generalized-Mean pooling layer.
Using GeM as a feature extractor achieves the current state of the art on the data sets that we will use for our experiments, Oxford5k~\cite{philbin2007object}
and Paris6k~\cite{Philbin08}, which are described in more detail below.
We chose to use the structure of ResNet-101.
% GeM replaces the fully-connected layer of ResNet-101 with a Generalized-Mean pooling layer.
GeM discards the fully-connected layer and replaces the average pooling layer of ResNet-101 with a Generalized-Mean pooling layer.
%We established that this wasn't true
%Note that technically a slightly higher performance could have been achieved with VGG rather than ResNet-101, but the difference is inconsequential for our experiments.
The model is pre-trained on ImageNet and fine-tuned using a data set that consists of 120k Flickr images provided in ~\cite{Radenovic-TPAMI18} following a structure-from-motion (SfM) pipeline.
The fine-tuning data set is a subset the data set of~\cite{schonberger2015single}, which contains 7.4 million images from Flickr with keywords of landmarks, cities and countries across the world.
The subset excludes Oxford and Paris.

GeM has been shown to outperform previous state-of-the-art approaches, which we mention here briefly for completeness. 
First,~\cite{sharif2014cnn} used off-the-shelf neural networks.
To improve retrieval performance, Neural Codes~\cite{babenko2014neural} used a fine-tuned CNN-model for neural feature extraction.
Finally,~\cite{tolias2016rmac} proposed regional maximum activation of convolutions (R-MAC) to improve image retrieval by adding an additional pooling layer to CNN-model.

%Due to its performance in benchmark data sets, we use GeM as neural feature extractor to build neural- feature-based CBIR.

%Check this sentence
The representations that are used by local-feature-based CBIR systems are generally Bag-of-Visual-Word (BoVW) models, dating back to~\cite{sivic2003video}.
%Mostly, local-feature-based CBIR systems relied on Bag-of-Word model (BoW)~\cite{sivic2003video}.
Codebooks containing a certain number of visual words are trained on extracted SIFT features.
%Codebooks of certain length are trained on extracted SIFT features.
We adopt a classic BoVW model with Hamming Embedding (HE)~\cite{jegou2008hamming}, which provides binary signatures that refine the visual-word-based matching.
%We use SIFT features with Hamming Embedding to represent the local feature based-CBIR.
Following~\cite{jegou2008hamming}, we extract SIFT feature of images and train codebooks of size 20,000. 
%Binary signatures of length 64 are used in HE setting, and matching appropriate threshold is set to 24.
Binary signatures of length 64 are used in the HE setting, and the threshold is set to 24.
Note that the basic BoVW system that we adopt performs competitively with approaches such as VLAD in~\cite{jegou2010aggregating}. 
For this reason, we are confident that it meets the needs of the experiments we perform here. 
We save more detailed investigation of techniques such as geometric matching and query expansion for future work.

% VLAD
%VLAD~\cite{jegou2010aggregating} increased the search efficiency and kept a comparable accuracy to previous methods.
% ASMK
% ~\cite{TAJ13} achieved state-of-the-art result by combining feature aggregation and matching techniques.

The representations used by a global-feature-based CBIR system capture information about overall image texture and image color, rather than information about specific keypoints.
Color histograms and Edge histograms (MPEG-7 descriptors) are commonly used for extracting global features.
For our experiments we adopt two widely-used global feature representations:
Color and Edge Directivity Descriptor (CEDD)~\cite{chatzichristofis2008cedd}, which combines image color and texture information, and
GIST~\cite{oliva2001modeling}, which extracts a holistic image representation reflecting the shape of a scene.
%We use CEDD and GIST features to construct the global feature based-CBIR.
% global details

%Check to see if this is correct.
We perform experiments with two types of image queries: whole image queries (designated WI) and bounding box queries (designated BB). 
The BB queries use only the content of a bounding box that focuses on the main subject matter of the image. 
This bounding box is pre-defined (it is included with the queries in the data sets).
We use BB queries in order to make our work comparable to other papers who test on the same data sets.
%Whole image (WI) queries are use to make sure that adversarial examples generated by PIRE and KR\&I can also be tested in global feature-based CBIR.
%Except that, bounding box (BB) queries are also used in neural-feature-based CBIR and local-feature-based CBIR such that our results are comparable to related works.
%The bounding box query represents an object by selecting a region of a query image in Oxford5k and Paris6k data sets.

To evaluate, we compare the quality of the results list returned using the original image as a query with the results list returned by adversarial query (i.e., the modified image). 
We adopt mean average precision (mAP), a standard information retrieval evaluation metric, to measure results list quality.
An image modification approach is successful if we observe a decrease in mAP when we move from the original query to the modified query.
%The decrease of mAP from original to adversarial queries represents the performance of PIRE and KR\&I.
Finally, to evaluate visual quality, we use
structural similarity (SSIM)~\cite{wang2004image}, which assesses the degradation of structural information to be presumed related to the human-perceived quality.
% The framework of this paper is shown in Figure~\ref{fig:sys-flow}.
% % what is CBIR 
% % different representations
% % explain GeM
% % what it is 
% % how it is implemented

\subsection{Data}
\label{data}
% standard statistics about the datasets.
We perform our experiments on two data sets: Oxford5k and Paris6k, which are publicly available and widely used in CBIR research.
Because so much work has been done on these data sets, what constitutes state-of-the-art performance is well understood, and we can be certain that when we test the blocking effects of adversarial queries on CBIR, we are testing a strong CBIR system.
%To make our results comparable to related works in CBIR, Oxford5k and Paris6k benchmark datasets are used in our experiment.
The Oxford5k data set consists of 5063 images and includes 55 standard queries representing different views/parts of 11 Oxford buildings.
Paris6k data set consists of 6412 images and also includes 55 standard queries from 11 different Paris landmarks.
Both data sets include distractor images, which are not related to any of the queries in the data set.
%Images in both data sets are assigned one of four possible labels, i.e., Good, OK, Junk and Absent.
%In retrieval performance evaluation, Good and OK are regarded as positive examples. 
%Absent images and Junk images are regarded as negative null examples respectively.
% For each landmark, the number of relevant images ranges from 7 to 289.

\section{Neural-Feature-based CBIR}
\label{Neural_CBIR}

In this section, we propose a simple yet effective algorithm, Perturbations for Image Retrieval Error (PIRE), which blocks neural-feature-based CBIR by perturbing pixels of the image query.

\subsection{Adversarial Queries with PIRE}
\label{PIRE}

%Specifically, we modify the query in a way that the resulted features can be pushed away from the original one. 
The basic innovation of PIRE is to modify the original image by pushing its feature representation away from the original position in feature space.
Specifically, PIRE maximizes the distance between the feature representation of the original image and that of the modified image, while at the same time limiting the overall size of the permutation.
Recall that PIRE is designed with the assumption
%that the information of the neural feature extractor model 
that the CNN-model (GeM~\cite{Radenovic-TPAMI18} in this paper) is available, and it aims to modify the input image with perturbations that are barely perceptible to the human eye.

PIRE is presented in Algorithm~\ref{alg:T1}.
$\boldsymbol{x}$ represents the image query, and $\boldsymbol{v}$ represents the perturbation vector.
We start with a random perturbation feature vector that has the same size of the image and update it by optimizing the following objective function:
\begin{align}\label{eq:loss}
    \text{maximize}\quad  &\| f(\boldsymbol{x}) - f(\boldsymbol{x} + \boldsymbol{v})\|_2^2  \\
    \text{subject to} \quad &\|\boldsymbol{v}\|_{\infty} \leq \epsilon\nonumber 
\end{align}

This optimization process will stop when the iterative conditions are met. 
We create the final perturbation using a multiplicative factor, here, set to 10, to guarantee that the perturbations are retained when the image is saved in an 8-bit format. 
We return to address this factor in more detail in Section~\ref{sec:save}.
% \section{Neural-Feature-based CBIR}
% In this section, we propose a naive algorithm Perturbations for Image Retrieval Error (PIRE) to perturbing all images which defeat CBIR using neural image representations.
% \subsection{Adversarial Queries with PIRE}
% We propose a method based on adversarial examples techniques to modify the extracted neural feature in order to cause image retrieval error.

% In PIRE, it is assumed that the information of the neural feature extractor model is available, and it aims to modify the image representation while keeping the image appeal.
% Specifically, we propose to calculate perturbation vectors that enlarge the distance between the protected neural feature vector and the original neural feature vector.

% PIRE starts with a random perturbation vector that has the same size as the image query.
% We define a loss function that measures the Euclidean distance between the original image feature and perturbed image feature.
% During the optimization process, the perturbation vector is updated to maximize this loss.
% This optimization process keeps running until iterative conditions are met.
% To avoid the perturbation vector vanish when saving as a compressed format, we enlarge 10 times the perturbation vector.
% Adding generated perturbation vectors to the original image strongly influences neural-feature-based CBIR system.

\begin{algorithm}
    \SetKwInOut{Input}{Input}
    \SetKwInOut{Output}{Output}

    \Input{Image query $\boldsymbol{x}$; Neural feature function $f$; Iteration limit $T$; Perturbation vector range $\epsilon$;}
    \Output{Adversarial image query $\boldsymbol{x} + 10 * \boldsymbol{v_{i}}$;}
    
    $w, h = \text{size}(\boldsymbol{x}_{1})$;\\
    $i = 1$;\\
    Generate a random matrix $\boldsymbol{v}_{0_{(w \times h)}}$;\\
    
    \While{$i < T$}{
    
    Calculate the distance between original image $\boldsymbol{x}$ and perturbed image  $\boldsymbol{x} + \boldsymbol{v}_{i - 1}$;
    \begin{flalign*}\hspace{-1.7cm}
        \boldsymbol{v_{i}} &= \argmaxl_{\boldsymbol{v}} \| (f(\boldsymbol{x}) - f(\boldsymbol{x} + \boldsymbol{v}_{i - 1}))\|_2^2;   
    \end{flalign*}\\
    Project $\boldsymbol{v_{i}}$ into a $L_{\infty}$ norm sphere;
    \begin{flalign*}\hspace{-2.9cm}
    \boldsymbol{v_{i}} &= \text{clip} (\boldsymbol{v}_{i}, -\epsilon, \epsilon);
    \end{flalign*}\\

        $i = i + 1$;
}
    Return perturbed image query;
    \begin{flalign*}\hspace{-4.2cm}
    \textbf{return} \quad \boldsymbol{x} + 10 * \boldsymbol{v_{i}};
    \end{flalign*}
    \caption{Perturbations for Image Retrieval Error (PIRE)}
    \label{alg:T1}
\vspace{-0.3cm}
\end{algorithm}

In each iteration, the perturbation vector is updated using the Adam optimization algorithm~\cite{kingma2014adam}.
% After iterating $T$ rounds, the distance of original image query and the resulting perturbed image will be a real number $d\in [0, \sqrt{2}]$.
In our experiments we look at the impact of T, the number of rounds iterated.
When the iterative conditions are met, perturbation vector $\boldsymbol{v_{i}}$ is the calculated perturbation vector.

In order to test PIRE, we apply it to all the query images of our data sets to create adversarial images, which are then saved.
Table~\ref{tbl:nf-map-org} reports results for the original queries and for adversarial queries created with PIRE (T = 500). 
It can be seen that the mAP drops dramatically, indicating that PIRE is highly successful.
Note that here we report bounding box (BB) queries only, and we are not yet concerned with the visual appearance of queries.
As we will in Section~\ref{sec:Adv_practice}, the choice of T allows us to control the trade off between PIRE's adversarial effect and its visual impact.
%for bounding box (BB) part of image queries in  
%It is concluded that PIRE (T = 500) strongly decreases the performance of neural-feature-based CBIR.
%In summary, given the query images of data sets and neural feature extractor, we calculate adversarial query images by PIRE.
%After the modification, we save perturbed image query and calculate the mAP and SSIM.
%Following the experimental protocol in Section~\ref{sec:exp_fra}, we show the performance evaluation results of PIRE (T = 500) for bounding box (BB) part of image queries in Table~\ref{tbl:nf-map-org}. 
%It is concluded that PIRE (T = 500) strongly decreases the performance of neural-feature-based CBIR.

\begin{table}[]
\newcommand{\tabincell}[2]{\begin{tabular}{@{}#1@{}}#2\end{tabular}}
\caption{Performance (mAP) of neural-feature-based CBIR (GeM~\cite{Radenovic-TPAMI18}) on Oxford5k and Paris6k data sets before and after original PIRE ($10*v$) modification with T=500 iterations.}
\vspace{-0.2cm}
\begin{tabular}{lcc}
\toprule
& \tabincell{c}{Oxford5k \\(BB)}               & \tabincell{c}{Paris6k \\ (BB)}             \\
\midrule
Original       & 78.39      & 87.27         \\
PIRE (T = 500) & 5.51       & 9.34  \\
\bottomrule
\end{tabular}
\label{tbl:nf-map-org}
\vspace{-0.3cm}
\end{table}

\subsection{Adversarial Queries in Practice}
\label{sec:Adv_practice}
%We now turn to discussion of the practical aspects of PIRE.
\subsubsection{Saving queries}
\label{sec:save}
% To avoid the perturbation vector vanish when saving as a compressed format, we add 10 times of generated perturbation to original image.
In order for PIRE to be used in practice, it is necessary that adversarial queries remain adversarial when they are saved.
When saving an image in JPEG format (uint8), float values that do not fit into 8 bits are approximated.
This means that the perturbations that PIRE adds to an image should not be so small that they disappear when the image is saved.
In ~\cite{carlini2017towards}, a method based on greed search was proposed to avoid the rounding effects discussed above. 
Saving images is obviously important, and so in Algorithm~\ref{alg:T1}, on the last line, we use a multiplicative factor 10 in order to make sure that our perturbations survive rounding. 
% Note that we find saving images obviously important, although much work on adversarial examples for image classifiers does not address it (cf.~\cite{Moosavi-Dezfooli_2017_CVPR}).\textcolor{green}{~\cite{carlini2017towards}}
%XXX If we can add (cf.~\cite{}) it would be great.

%From experiment, we found that if saving the perturbed image with adding one time $\boldsymbol{v}$, the influence of perturbation may vanish.
%We reason that this phenomenon is caused by image format.
%As a consequence, perturbation vector is also approximated, and it may not have influence after saving image.
%To apply the generated perturbation vector, we get the image data and add 10 times of the generated perturbation vector on it.
However, this approach of blindly making perturbations large is not elegant, since large perturbations lead to artifacts that are visually obvious.
To tackle this issue, we propose a refinement to PIRE.
The refinement adds a function
%XXX check this, I assume p and not g was meant
$p(\boldsymbol{v_{i}})$ that magnifies the original perturbation vector to be just large enough not to be rounded away when the image is saved. 
These controlled perturbations improve the visual appearance of the adversarial queries.
As the final step, the refined PIRE algorithm returns $\boldsymbol{x}_{q} + p(\boldsymbol{v}_i)$ as the modified image query.
%that is able to block the neural-feature-based CBIR.
Exploratory experiments allowed us to observe that refined PIRE is able to substantially reduce the amount of perturbation needed to achieve an adversarial effect, and, as will be discussed below, also improves the visual appearance. 
We point out that the example in Figure~\ref{fig:dl-ex1} is an actual query from the Oxford5k data set tested with respect to our neural-feature-based CBIR system. The original query image achieves an AP of 93.95 and for the adversarial query (created with refined PIRE T=500) the AP plunges to 3.77.
The PIRE results in the rest of the paper are for refined PIRE.
%Results of the refined PIRE (T=500) are presented in Table~\ref{tbl:nf-map}. 
%Recall that the mAP for adversarial queries (BB) was 5.51 for Oxford5k and 9.34 for Paris6k. With refined PIRE (BB), these values have dropped to 3.93  
%sufficiently so that they survive being saved.
%$p$ mainly enlarges the original perturbation vector to the level that the values are larger than a 8-bit, which makes sure that saving in JPEG will not lose the perturbation information and keeping the image perception.

\subsubsection{Viewing queries}
\label{sec:view}
%XXX I don't understand the reasoning in these next two sentences.
%By applying PIRE, it is possible to control the distance between original image query feature and adversarial image query feature.
%As a consequence, we do experiments to investigate the influence of iterative rounds (perturbation level) on mAP and SSIM.
Next, we focus on the experience of users viewing adversarial queries.
In order to get further insight into the visual impact of PIRE perturbations, we experimented with different levels of perturbation.
Specifically, we prepared adversarial image queries with refined PIRE for two different representative values of the threshold T, which controls the number of iterations used to calculate the perturbations.
Table~\ref{tbl:nf-map} shows that adversarial image query generated with fewer rounds (T=200) still strongly decreases the performance of neural-feature-based CBIR.

\begin{table}[]
\newcommand{\tabincell}[2]{\begin{tabular}{@{}#1@{}}#2\end{tabular}}
\caption{Performance (mAP) of neural-feature-based CBIR (GeM~\cite{Radenovic-TPAMI18}) on Oxford5k and Paris6k data sets before and after query modification by PIRE ($p(\boldsymbol{v})$)  (T=200 and T=500).}
% \vspace{-0.3cm}
\begin{tabular}{lcc}
\toprule
& \tabincell{c}{Oxford5k \\(BB / WI)}               & \tabincell{c}{Paris6k \\ (BB / WI)}             \\
\midrule
Original       & 78.39/74.42          & 87.27/87.26         \\
PIRE (T = 200) & 22.98/18.00          & 34.49/26.53         \\
PIRE (T = 500) & 3.93/2.31           & 10.53/7.18         \\
% Gaussian Noise & 72.90/71.45           & 84.21/85.64         \\
\bottomrule
\end{tabular}
\label{tbl:nf-map}
\vspace{-0.2cm}
\end{table}

\begin{figure}[htb!]
\includegraphics[width=1.0\columnwidth]{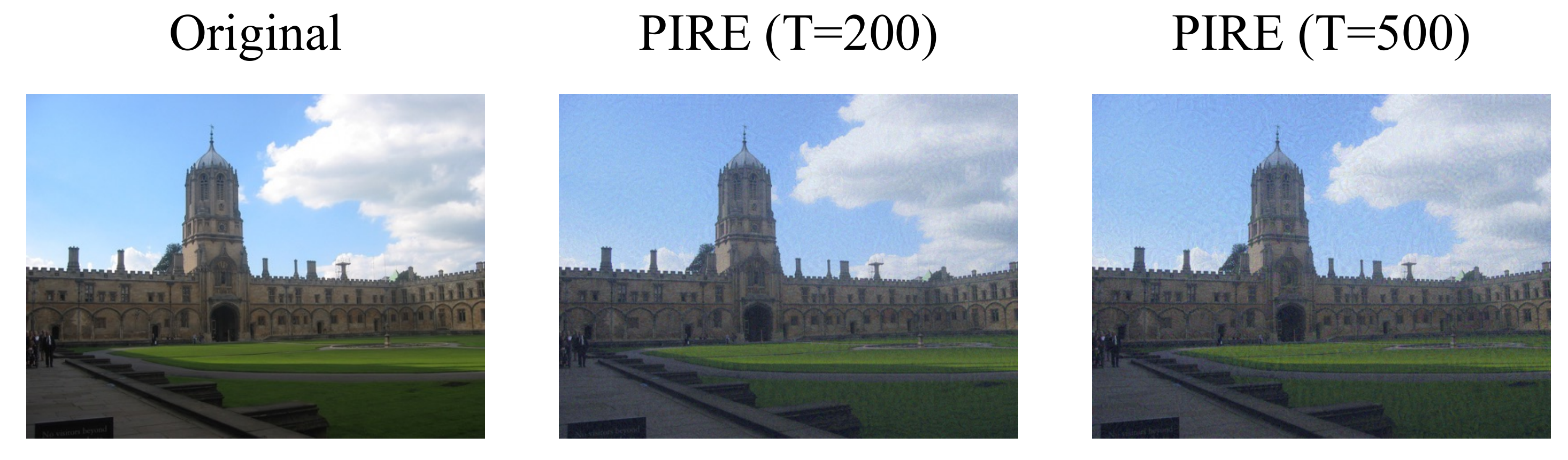}
%\vspace{-0.5cm}
\caption{Examples of original image queries vs. adversarial queries generated using PIRE with different number of iterations (T=200 and T=500). Fewer iterations lead to less visible perturbations. (Best viewed on screen with magnification.)}
\label{fig:NF-imgs}
\end{figure}

In Figure~\ref{fig:NF-imgs}, we present example queries to illustrate the contrast between PIRE using different values of the threshold on the number of iterations (T=200 and T=500). 
%In Oxford5k data set, the mAP of bounding box queries drops from 78.39 to 3.93.
%In Paris6k data set, the mAP of bounding box queries drops from 87.27 to 10.53.
In order to quantify the relative difference in impact on the visual appearance, we report SSIM values in Table~\ref{tbl:nf-ssim}. 
%shows that PIRE maintains the perceived quality of adversarial queries, and less iterative rounds increase the visual quality of adversarial image queries.
Although more iterations lower the SSIM, the quality is still acceptable at both levels.
%XXX Did we have a paper that mentioned what hte level is.
In addition, we compared the SSIM value of the original PIRE ($10*v$) and the refined PIRE ($p(\boldsymbol{v})$).
%%ML+ Check this: I think it should be average SSIM values and not "results"
SSIM on BB queries from Oxford5k (T = 200) went from 0.757 (PIRE) to 0.801 (refined PIRE). 
For BB queries on Paris6k (T = 200) results went from 0.690 (PIRE) to 0.771 (refined PIRE).

\begin{table}[]
\newcommand{\tabincell}[2]{\begin{tabular}{@{}#1@{}}#2\end{tabular}}
\caption{Average image quality (SSIM values) of adversarial queries from Oxford5k and Paris6K data sets generated by PIRE. (The SSIM value of the original query equals 1; BB= Bounding Box and
WI=Whole Image.)}

\begin{tabular}{lcc}
\toprule
& \tabincell{c}{Oxford5k \\(BB / WI)}               & \tabincell{c}{Paris6k \\ (BB / WI)}             \\
\midrule
PIRE (T=200)   & \textbf{0.801}/\textbf{0.754}         & \textbf{0.793}/\textbf{0.771}               \\
PIRE (T=500)  & 0.738/0.687        & 0.727/0.716             \\

\bottomrule
\end{tabular}
\label{tbl:nf-ssim}
\vspace{-0.2cm}
\end{table}

\subsubsection{Protecting queries}
Next, we demonstrate that PIRE has potential to cause a drop in the mAP of a CBIR system when the neural network used for indexing is unknown.
We used a new neural network architecture, VGG-GeM, as the basis for generating adversarial queries with PIRE.
We tested these queries against our original neural-feature-based CBIR system, which uses ResNet-GeM.
We observed a mAP drop from 74.42 to 2.91 (Oxford5k), and from 87.26 to 9.39 (Paris6k) (T=500; cf. Table~\ref{tbl:nf-map}). 
We note that when VGG-GeM is used for both PIRE and retrieval, the effect is comparable to when ResNet-GeM is used for both PIRE and retrieval.
We do not investigate VGG-GeM in more detail here, since the SSIM is ca. 0.15 lower for PIRE queries created with VGG-GeM than for PIRE queries created with ResNet-GeM.
%Then, we reversed the role of the architectures.
%We found that query image perturbations optimized with respect to a ResNet-GeM architecture cause a mAP drop with respect to a VGG-GeM-feature-based CBIR system of 77.57 to 42.00 (Oxford5k), and from 82.67 to 64.52 (Paris6k). 
%We leave discussion of the source of the relative size of the drops in the two cases to future work.
Our conclusion here is that PIRE has the potential to lower mAP without access to information on the architecture used for indexing.
This conclusion is consistent with a further set of exploratory CBIR experiments we carried out with Google Images (\url{https://images.google.com}).
The details of this system are unknown to us, but we assume that advanced neural representations are used, and that the background collection (index) is very large.
We found the existence of a unexpectedly high number of cases in which the results returned by Google Images are impacted by PIRE.
Future work on the investigation of nature of this impact promises to yield further interesting insight.

\subsubsection{Editing Queries}
\label{sec:edit}

Simple image transformations, such as resizing and cropping, may destroy the specific structure of adversarial perturbations.
This effect was pointed out by~\cite{xie2018mitigating} for image classification.
%%XXX do we know why we used only Oxford?
Here, we use the Oxford5k data set to test the robustness of the perturbations generated by our PIRE against resizing and cropping of the adversarial query. 
These transformations obviously will also affect the performance of the original queries, so we report results for those as well.
%Since the performance of CBIR with original query is also be affected by image cropping and resizing, we report results for both origiPIRE along with the original performance in each case.
\begin{table}[]

\caption{Performance (mAP) of neural-feature-based CBIR (GeM~\cite{Radenovic-TPAMI18}) on the Oxford5k data set with bounding box queries and different resizing/cropping settings. Original image queries are compared with PIRE adversarial queries.}
\vspace{-0.4cm}
\begin{tabular}{lcccccc}
\\ \toprule
 Resizing&50\% &80\%& 100\%&  150\% &200\%
\\ \midrule
Original &65.09 &74.31 & 78.39& 71.30 &  62.03       \\ 
PIRE (T = 500)  &55.91&  41.41 & 3.93 &16.08  &12.06  
\\ \toprule
                  Cropping&100\%& 90\%&80\% & 60\% & 40\%   \\ \midrule
Original &  78.39& 76.01&76.01 &69.54 &46.89              \\ 
PIRE (T = 500) &3.93 & 24.13& 27.26 & 25.81 &10.35                \\ 

\bottomrule
\end{tabular}
\label{tbl:resize_crop}
\end{table}

For image resizing, we implement upscaling and downscaling operations, resulting in resized image queries with 200\%, 150\%, 80\% and 50\% of the original size.
From the results, which are reported in Table~\ref{tbl:resize_crop}, it can be observed that upscaling has only a small influence on PIRE (i.e., PIRE mAP remains lower than original mAP), while downscaling has larger influence (i.e., PIRE mAP and original mAP are closer).
We suspect that the effect of adversarial queries lies in the subtle perturbation of the original pixels, downscaling changes most of the perturbed pixels. 
On the other hand, upscaling only interpolates new pixels between the perturbed pixels and for this reason does not impact the structure of perturbation as strongly. 

For image cropping, we apply four different settings, i.e., 40\%, 60\%, 80\% and 90\% of the original size.
From the results, which are also reported in Table~\ref{tbl:resize_crop}, we can observe that image cropping has more impact than resizing on the original performance of CBIR, which we attribute to the loss of image content. 
However, PIRE remains effective, and causes substantial performance drops in different settings of image cropping.

% \begin{table}[]
% \caption{Performance (mAP) of neural-feature-based CBIR (GeM~\cite{Radenovic-TPAMI18}) on the Oxford5k data set with bounding box queries and different cropping settings. Original image queries are compared with PIRE adversarial queries.}
% \begin{tabular}{lccccc}
% \\ \toprule
%                   &100\%& 90\%&80\% & 60\% & 40\%   \\ \midrule
% Original &  78.39& 76.01&76.01 &69.54 &46.89              \\ 
% PIRE (T = 500) &3.93 & 24.13& 27.26 & 25.81 &10.35                \\ 

% \bottomrule
% \end{tabular}
% \label{tbl:crop}
% \end{table}

\subsubsection{Leaking queries}
\label{sec:leak}

% attacker know technique-1, it would not improve the situation.
If PIRE is used in practice, it can be expected that some images that have been perturbed with PIRE find their way (i.e., ``leak") into the background collection.
We use one query from each of our data sets (christ-church-4 for Oxford5k and triomphe-3 for Paris6k) to explore what happens when not only queries, but also background images are perturbed with PIRE.
%To evaluate the robustness of PIRE, it is considered the case that attacker already had information of PIRE.
%To increase the performance of neural-feature-based CBIR, they may use the same technique and add perturbed images to background dataset in order to improve the CBIR system against PIRE.
For each query, we replace all its original relevant images (i.e., the ones labeled \emph{good} or \emph{ok}) in the background collection with adversarial versions using PIRE (T=200). 
We test two cases: one in which the image queries are perturbed with exactly the same setting of PIRE (T=200), and one in which they are perturbed with a different setting (T=500).

\begin{table}[]
\caption{Impact of PIRE on the performance (AP) of neural-feature-based CBIR (GeM~\cite{Radenovic-TPAMI18}) for two specific queries before and after replacing the relevant images for these queries in the background collection.}
\vspace{-0.5cm}
\[
\begin{tabular}{lccc}
\\ \toprule
         Background   &  Query   & christ-church-4 & triomphe-3 \\ \midrule
Original   &   Original    &93.88         &89.52               \\ 
Original & PIRE (T=200)          &2.83         &7.39               \\ 
Replaced & PIRE (T=200)     &34.52          &48.85               \\
Replaced & PIRE (T=500)     &22.43          &36.42               \\

\bottomrule
\end{tabular}
\]
\label{tbl:nf-bkpt}
\vspace{-0.5cm}
\end{table}

The results, reported in Table~\ref{tbl:nf-bkpt}, demonstrate that adversarial queries can still maintain an adversarial effect when the relevant background images have been perturbed. 
If the relevant background images are perturbed with a different T than the adversarial query, the adversarial effect is stronger (mAP is lower) than when they are perturbed with the same T.
These results suggest that adversarial queries leaking into the background collection might diminish, but will not negate, the adversarial effect over time.
An approach to maintaining the strength of the adversarial effect would be to promote the use of diverse perturbation settings.

%From the results in Table~\ref{tbl:nf-bkpt}, we can observe that the replacement of background images increases the performance of neural-feature-based CBIR against PIRE with a relatively small extent.
%Replacing the background images with the same setting (PIRE=200) yields more improvement than different setting.

%These results verify the robustness of our PIRE approach. We suspect that by generating perturbations in an unsupervised way, PIRE pushes the neural features of all images to a certain area in the feature space, where the non-relevant images representing a certain pattern (as shown in Figure~\ref{fig:dl-ex1}) causes the performance drop of the CBIR system.

% \subsection{Discussion and Conclusion}
% \label{sec:dis_con}

% Better explanation of the novelty of perturbing the layer that we perturb
% Better understanding of the ‘mutual awareness’ between PIRE and the network used to create the neural feature representations.
% Influence of the initial random perturbation vector

% We need to study these points more

% protected image every time different  
% protected images and accumulated in facebook

% resizing/cropping ruins the effect (future work) ---> test it
% what if attacker is attacking with local feature-based CBIR.
% The feature space and distance

% \section{The power of local and global features}
\section{CBIR beyond Neural Features}
\label{sec:CBIR_beyond}

% if we are worried about protection of images, more worried a task with a local feature-based CBIR system.
% \subsection{Power of local features to attack}
%In local-feature based CBIR, KR\&I modification can be applied to generate adversarial queries.
%In this section, we test PIRE and different KR\&I-modifications with local-feature-based CBIR, and we test adversarial queries generated by PIRE and KR\&I-modification in global-feature based CBIR system.

In order to understand the larger implications of adversarial queries, we now turn to look at local and global image features.

\subsection{Local-feature-based CBIR}
\label{sec:CBIR_SIFT}
Here, we test PIRE against the SIFT-based CBIR system introduced in Section~\ref{CBIR_system}, and compare it to existing KR\&I-modifications that have been developed to block retrieval with local features. 

\subsubsection{KR\&I-modification}
\label{sec:KRI}
We use the central methods from previous work to remove and inject SIFT keypoints.
From~\cite{do2010deluding}, we test Removal with Minimum local Distortion (RMD) and Local Smoothing (LS), as well as the Forge new keypoints with Minimum local Distortion (FMD) method, which is representative of keypoint injection.
In addition, we also test Removal via Directed Graph Construction (RDG)~\cite{li2017sift}, a SIFT keypoint removal method that explicitly addresses visual quality. 

\begin{figure}
% \vspace{-0.3cm}
\includegraphics[width=1.0\columnwidth]{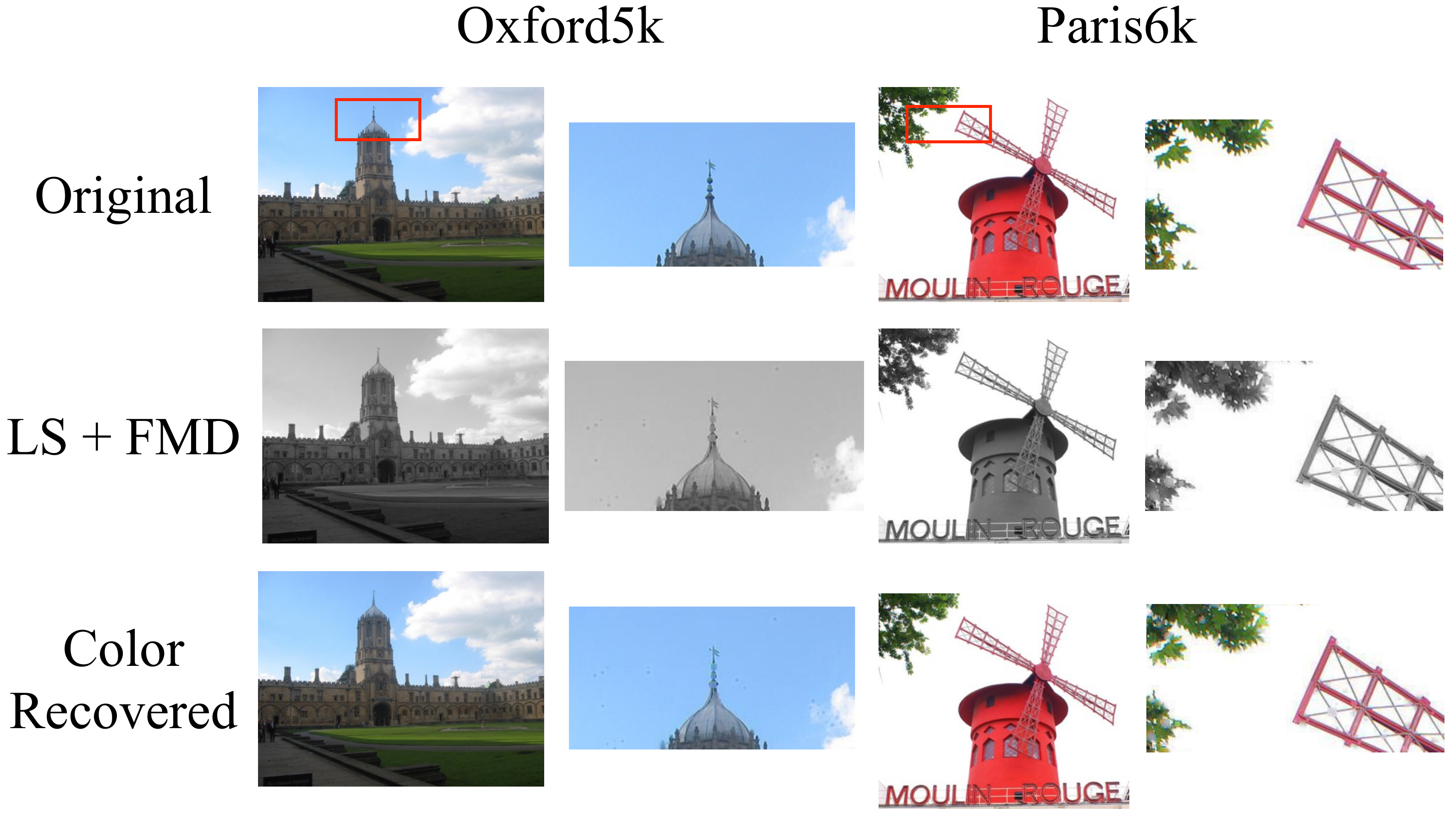}
\vspace{-0.5cm}
\caption{Examples of SIFT KR\&I: original image queries with specific-region enlargements (top row), gray-scale images modified with LS + FMD (middle row), and the color recovered version of the modified images (bottom row).}
\label{fig:lf-sift}
%\vspace{-0.5cm}
\end{figure}

We carry out experiments with SIFT-based CBIR on the original queries, on queries modified with five different KR\&I methods, and on adversarial queries created with PIRE. 
Results are presented in Table~\ref{tbl:lf-siftKPRI}. 
Only two KR\&I methods (RMD + LS and LS + FMD) achieve substantial success in lowering the mAP compared to the mAP of the original queries, and some increase it (by unintentionally streamlining the visual word vocabulary).
Interestingly, PIRE (T=500), although it is designed to be adversarial with respect to neural-feature-based CBIR, shows a blocking effect with respect to SIFT-based CBIR. 
For the Oxford5k data set, this effect is on par with the best of KR\&I methods.
We point out that KR\&I methods maintain a better image quality than PIRE, as can be seen from Table~\ref{tbl:lf-ssim}, which reports SSIM for RMD + LS and LS + FMD.

\begin{table}[]
\newcommand{\tabincell}[2]{\begin{tabular}{@{}#1@{}}#2\end{tabular}}
% \vspace{-0.5cm}
\caption{Performance (mAP) of SIFT-based CBIR on Oxford5k and Paris6k data sets: original queries and after modification with KR\&I and PIRE. (BB=Bounding Box and
WI=Whole Image.)}

\begin{tabular}{lcc}
\toprule
& \tabincell{c}{Oxford5k \\(BB / WI)}               & \tabincell{c}{Paris6k \\ (BB / WI)}             \\
\midrule
Original &  52.57/51.59     &  45.46/44.63                \\
RDG  &  53.00/51.08         &  44.45/44.44              \\
RMD   &  53.81/53.02        &  44.47/45.23               \\
RMD + LS  &  42.54/46.90   &  32.75/33.60              \\
FMD  &  54.20/51.77         &  42.38/44.58              \\
LS + FMD  &  41.23/\textbf{42.21}    &  \textbf{29.86}/\textbf{32.64}            \\
PIRE (T = 500)  & \textbf{40.90}/44.05     & 39.23/40.73              \\
\bottomrule
\end{tabular}
\label{tbl:lf-siftKPRI}
\end{table}
% \begin{table}[]
% \newcommand{\tabincell}[2]{\begin{tabular}{@{}#1@{}}#2\end{tabular}}
% \caption{Performance (mAP) of SIFT-based CBIR on Oxford5k before and after  KR\&I-modification and PIRE. (BB= Bounding Box and
% WI=Whole Image)}

% \begin{tabular}{cccccccc}
% \toprule
% &Original&RDG&RMD&\tabincell{c}{RMD \\+\\ LS}&FMD&\tabincell{c}{LS \\+\\ FMD}&\tabincell{c}{PIRE \\(T = 500)}\\
% \midrule
% BB &  52.57  &  53.00&  53.81 &   42.54&  54.20    &  41.23&  \textbf{40.90}      \\
% WI &51.59 &51.08&53.02& 46.90 &51.77  &\textbf{42.21}  &44.05\\    
% \bottomrule
% \end{tabular}
% \label{tbl:lf-siftKPRI}
% \end{table}

% \begin{tabular}{lcc}
% \toprule
% & BB & WI               \\
% \midrule
% Original &  52.57&51.59          \\
% RDG  &  53.00&51.08                      \\
% RMD   &  53.81&53.02          \\
% RMD + LS  &  42.54& 46.90                      \\
% FMD  &  54.20&51.77              \\
% LS + FMD  &  41.23&42.21                       \\

% PIRE (T = 500)  &  40.90&44.05              \\
% \bottomrule
% \end{tabular}
% \label{tbl:lf-siftKPRI}
% \end{table}

% \begin{tabular}{lcc}
% \toprule
% & \tabincell{c}{Oxford5k \\(BB / WI)}               \\
% \midrule
% Original &  52.57/51.59          \\
% RDG  &  53.00/51.08                      \\
% RMD   &  53.81/53.02          \\
% RMD + LS  &  42.54/ 46.90                      \\
% FMD  &  54.20/51.77              \\
% LS + FMD  &  41.23/42.21                       \\

% PIRE (T = 500)  &  40.90/44.05              \\
% \bottomrule
% \end{tabular}
% \label{tbl:lf-siftKPRI}
% \end{table}

\subsubsection{SIFT color recovery}
Since, essentially, SIFT features are extracted from single-channel images, in general, KR\&I-modifications can only be applied to gray-scale images.
However, because we are interested in visual appearance, we would like to compare color versions of KR\&I-modified images.
To this end, we propose a naive method to recover color after KR\&I modification.
Color recovery also allows us to make a fair comparison between the impact of PIRE and KR\&I modification on global-feature-based CBIR in Section~\ref{sec:global}.
%Moreover, color restoration is useful in the practical application where users of the social networks would not like to sacrifice the color of the original image for the modifications. 

Given an original color image $\boldsymbol{I}_{rgb}$ with three channels $\boldsymbol{I}_r$,$\boldsymbol{I}_g$ and $\boldsymbol{I}_b$, its gray-scale version $\boldsymbol{I}_{gray}$ can be calculated by the widely-used formula $\boldsymbol{I}_{gray}=0.30*\boldsymbol{I}_r+0.59*\boldsymbol{I}_g+0.11*\boldsymbol{I}_b$.
A successful recovery method should guarantee that the restored color image ${\boldsymbol{\hat{I}}_{rgb}}$ can be transformed back to the modified gray-scale image $\boldsymbol{I}_{mod}$ without the loss of modification effects, i.e., $\boldsymbol{I}_{mod}=0.30*{\boldsymbol{\hat{I}}_r}+0.59*{\boldsymbol{\hat{I}}_g}+0.11*{\boldsymbol{\hat{I}}_b}$.
In order to recover the color information, we multiply the pixel at each location (i,j) by the same ratio $\alpha$ for the three channels of the original image $\boldsymbol{I}_{rgb}$. The process can be formalized as
\begin{align*}
\boldsymbol{\alpha}(i,j)&=\boldsymbol{I}_{\text{mod}}(i,j)/\boldsymbol{I}_{\text{gray}}(i,j) \\  
\{{\boldsymbol{\hat{I}}_c}(i,j)|c\in\{r,g,b\}\}&=\boldsymbol{\alpha}(i,j)*\{\boldsymbol{I}_c(i,j)|c\in \{r,g,b\}\}
\end{align*}

Figure~\ref{fig:lf-sift} provides a impression of the image quality after modification with LS + FMD, using two example queries.
For each example, details in the red square are enlarged and shown alongside the whole image query.
Our simple color recovery method appears to achieve its aim well.
We can observe that artifacts are present in the gray-scale modified images. These are echoed in the color-recovered images. 
%It can be seen that the LS + FMD method blurred some images details and forged some new points, while RDG kept a high image quality.
%We test SSIM for five categories of modified images and show results in Table~\ref{tbl:lf-ssim}.
%The mAP results are reported in Table~\ref{tbl:lf-siftKPRI}.

\begin{table}[]
\newcommand{\tabincell}[2]{\begin{tabular}{@{}#1@{}}#2\end{tabular}}
\caption{Average image quality (SSIM values) of adversarial queries from Oxford5k and Paris6K data sets generated by KR\&I methods. (The SSIM value of the original query equals 1; BB= Bounding Box and
WI=Whole Image.)}

\begin{tabular}{lcc}
\toprule
& \tabincell{c}{Oxford5k \\(BB / WI)}               & \tabincell{c}{Paris6k \\ (BB / WI)}             \\
\midrule
LS + FMD   &  0.917/0.938          &  0.953/0.971                \\
RMD + LS   &  0.915/0.940         &  0.952/0.972              \\
\bottomrule
\end{tabular}
\label{tbl:lf-ssim}
\end{table}

% \subsection{Power of local features}

\subsection{Global-Feature-based CBIR}
\label{sec:global}
% intuition: protect global feature-based CBIR, we need to change images visibly.
% table shows that the best from local and neural above, do not impact global feature-based CBIR
%Potential approach to blocking global-feature-based CBIR by modifying the global features of the image query will inevitably change the image appearance visibly, resulting in limited use in the scenario of social multimedia.
Finally, we turn to investigating global-feature-based CBIR, using the CEDD and GIST systems described in Section~\ref{CBIR_system}.
%Therefore, we only test the impact of PIRE and the KR\&I method LS+PMD on the global- feature-based CBIR.
%As mentioned in~\ref{CBIR_system}, we carry out experiments with two well-known global features, namely CEDD and GIST.
CEDD is a low computational-cost feature that incorporates color and texture information in a histogram, while GIST features can represent perceptual dimensions (naturalness, openness, roughness, expansion and ruggedness) of a semantic scene by encoding coarsely localized information in the energy spectrum of an image~\cite{li2017sift}.

The results in Table~\ref{tbl:gf-ssim} reveal that image modifications operating on local features (LS + FMD) do not block global-feature-based CBIR.
However, our PIRE adversarial queries have a quite strong blocking effect on CEDD-based CBIR and a quite noticeable blocking effect on GIST-based CBIR.
These results are interesting since PIRE was not trained to block global-feature-based retrieval.
In order to understand their implications, we must know whether PIRE is acting specifically to disrupt pixel patterns that are important for global-feature-based CBIR, or it is merely acting as a sophisticated method of introducing noise throughout the image, which then has a blocking effect because it makes the overall image quality worse.
To this end, we carry out a baseline experiment using queries modified with  Gaussian noise.
Specifically, we generate noise for each image query such that the result is a SSIM value similar to the one caused by PIRE. (On average, for Oxford5k, SSIM equals 0.687 for PIRE, and 0.652 for Gaussian noise; for Paris6k, SSIM equals 0.716 for PIRE, and 0.708 for Gaussian noise.)

As shown in Table~\ref{tbl:gf-ssim}, Gaussian noise degrades performance in the case of GIST-based CBIR. 
We assume that the reason is that GIST extraction is based on spectral information, with which high-frequency noise interferes, and that PIRE is having a similar effect.

More interesting is how PIRE degrades the performance of CEDD-based CBIR.
Our explanation for these results is that CEDD captures texture information, and that the structural perturbations generated by PIRE interfere with texture more effectively than the random changes of Gaussian noise.
For completeness, we confirm that this effect does not account for the ability of PIRE to block neural-feature-based CBIR. In Table~\ref{tbl:nf-map}, we saw that PIRE (T = 500) drops the mAP of a neural-based CBIR system from 74.42 to 2.31 for the Oxford5k data set (Whole Image queries). 
Here, the effect of Gaussian noise contrasts with the effect of PIRE.
If Gaussian Noise instead of PIRE is used to modify the query, the drop is from 74.42 to only 71.45. 
Behavior on the Paris6k data set and with Bounding Box queries is comparable.

\begin{table}[]
\newcommand{\tabincell}[2]{\begin{tabular}{@{}#1@{}}#2\end{tabular}}
\caption{Performance (mAP) of global-feature-based CBIR (Whole image queries): original queries, PIRE (T=500), KR\&I modification, and Gaussian noise baseline.}

\begin{tabular}{lcc}
\toprule
& \tabincell{c}{Oxford5k \\(CEDD / GIST)}               & \tabincell{c}{Paris6k \\ (CEDD / GIST)}             \\
\midrule
  Original & 10.77/19.29          &  9.61/18.30               \\
   LS+FMD&  10.67/18.58         &  9.92/17.86              \\
   PIRE (T=500)& \textbf{2.54}/\textbf{14.71}        & \textbf{5.06}/12.11             \\
   Gaussian Noise&  10.68/\textbf{14.71}         &  8.27/\textbf{9.93}              \\
\bottomrule
\end{tabular}
\label{tbl:gf-ssim}
\vspace{-0.3 cm}
\end{table}

\section{Conclusion and Outlook}
\label{sec:conclusion}
This paper has made the case for studying adversarial queries in content-based image retrieval.
We have proposed a new algorithm called PIRE, which is a neural perturbation approach for creating adversarial queries. 
In contrast to previous work on adversarial examples, PIRE does not require supervision (i.e., no labels from the data set to which it is applied) and is for this reason suited for image retrieval scenarios.
Our experimental analysis of PIRE and of other, more traditional, approaches for blocking image matching with keypoint injection and removal (KR\&I) has provided valuable insight into adversarial queries.
We summarize these insights in terms of their implications for different groups of researchers.

\emph{Researchers in deep learning:} 
%PIRE is unsupervised, but beyond this special novelty it is otherwise a fairly standard neural perturbation approach. 
Our paper opens interesting topics in CBIR for researchers in deep learning. 
%This paper has shown the ability of PIRE, a relatively basic approach, to handily block neural-feature based CBIR. 
%Due to this finding, 
%we recommend, going forward, that research effort on adversarial examples in machine learning should move beyond image classification to include CBIR. 
%Our results suggest that adversarial queries can be constructed without complete knowledge of the network used to create the neural representations of the CBIR system that they block.
%Researchers in deep learning can now explore more sophisticated approach, and develop more sophisticated methods for
%An interesting area for future work is 
First, improvements on PIRE can be explicitly designed to generate queries that are adversarial with respect to a CBIR system for which little or no information is available, i.e., that uses arbitrary neural representations.
Next, we point out again that the data set used to fine-tune our CNN-model is semantically related to the data set to which PIRE is applied. 
Specifically, the fine-tuning data depicts buildings, but not specifically in our cities. 
In the future, the impact of this semantic relationship both on CBIR performance and on the ability of PIRE to block CBIR performance should be better understood.
We also point to~\cite{upir}, work on universal perturbations for image retrieval that came to our attention while preparing the camera-ready version of this paper.
Future work should further develop the ideas of~\cite{upir}, such as universal perturbations (PIRE is image specific) and pseudo-supervision.
%Finally, although we have focused on unsupervised adversarial queries here, pseudo-supervised methods are worth further investigation.

\emph{Researchers interested in local and global features:} 
%Our experiments have demonstrated that successful study of adversarial queries requires renewed research attention for traditional image features.
Our experimental analysis suggests that neural perturbations have potential to block local-feature-based CBIR and global-feature-based CBIR, opening interesting paths for future work.
Our results with a neural-feature-based CBIR system show that adversarial queries created with neural perturbations lose their blocking ability after certain edits. KR\&I approaches may not have these weakness.
%In contrast, queries created with keypoint injection and removal (KR\&I), retain their adversarial properties.
%Our experiments with a global-feature-based CBIR point to a direct link between visible changes to the query, and the ability of the query to block global-feature-based CBIR.
%Although global features may not provide the best CBIR performance, they are surprisingly robust in the face of adversarial queries.

\emph{Multimedia privacy researchers:} 
Not everyone who is able to deploy CBIR on a large collection of users' images can be expected to have the users' best interests in mind, and actors with ill intent are an inevitable risk.
Our results suggest that adversarial queries are a promising topic of study for multimedia privacy researchers.
%have good reason to continue research on adversarial examples with confidence that they have 
%can have confidence in the potential of adversarial approaches to provide protection against CBIR-based privacy violations.
Note that modest reductions in CBIR performance may already be enough to deincentivize malicious actors from abusing CBIR systems.
However, much research still lies ahead.
In order to implement privacy protection, it is necessary to apply modifications not only to query images, but also to images in the background collection.
%Our experiments have demonstrated that applying the same modification to both the query image and relevant background images leads to worse retrieval performance than when both the query and the background images are unmodified.
Our experimental results suggest that more work should be devoted to understanding the rate at which image modifications need to change dynamically in order to block CBIR over time.
Finally, in order for users to adopt image modifications to protect their privacy, it is necessary to pay close attention to the visual acceptability of the modified images.
Future work must focus both on minimizing the visual impact of perturbations, as well as understanding how to make visual changes that are acceptable to the user, in cases in which it is necessary to make visible changes.

In closing, we return to question in the paper's title: `Who's afraid of adversarial queries?'
Depending on the threat model that a researcher adopts, adversarial queries might be considered part of the problem or part of the solution, and at first consideration might be more or less scary. 
However, our overall answer is that no one should be afraid of adversarial queries, since they are important to understand and open up interesting new research questions. 
%With this paper we have shown the importance of adversarial queries, and have suggested fruitful directions for research to pursue next.
\vspace{-0.1 cm}

% \section{Introduction}
% % deep feature-based system has not caused massive jump performance.
% \section{Background}
% \subsection{}
% % 1. We are looking at whole image
% % 2. dataset, each section choose one that represents state-of-the-art.
% % state of the art technique and argue about experimental setup

% \section{Related work}

% % CBIR for the whole image.
% \section{Adversarial examples in neural feature-based CBIR} 
% \subsection{Approach}
% \subsection{Results}
% \subsubsection{Comparison between original and protected queries}
% % relative naive technique for creating adversarial queries will protect queries from neural feature-based CBIR.

% \subsubsection{Robustness of adversarial queries}
% % attacker know technique-1, it would not improve the situation.
% % protected image every time different  
% % protected images and accumulated in facebook

% % resizing/cropping ruins the effect (future work) ---> test it

% % what if attacker is attacking with local feature-based CBIR.

% \section{The power of local features}
% % if we are worried about protection of images, more worried a task with a local feature-based CBIR system.
% \subsection{Power of local features to attack}
% \subsection{Power of local features to protect}

% \section{The power of global feature-based CBIR}
% % intuition: protect global feature-based CBIR, we need to change images visibly.
% % table shows that the best from local and neural above, do not impact global feature-based CBIR
% \section{Conclusions}
% % Why we think the people have not done it before.

\bibliographystyle{ACM-Reference-Format}
\begin{acks}
This work is part of the Open Mind research program, financed by the Netherlands Organization for Scientific Research (NWO).
This work was carried out on the Dutch national e-infrastructure with the support of SURF Cooperative.
\end{acks}
\bibliography{sample-bibliography}

\end{document}